\documentclass[sigconf, screen, nonacm=true]{acmart}

\usepackage{booktabs}
\usepackage{multirow}

\title{LEDGER: A Long-Context Benchmark of Corporate Annual Reports for Grounded Financial Retrieval and Extraction}

\author{Charles Moslonka}
\orcid{0000-0002-8719-1510}
\affiliation{%
	\institution{Artefact Research Center}
	\city{Paris}
	\country{France}
}
\affiliation{%
	\institution{MICS, CentraleSupélec, Université Paris-Saclay}
	\city{Gif-sur-Yvette}
	\country{France}
}
\email{charles.moslonka@artefact.com}

\author{Amaury de Vitry}
\affiliation{%
	\institution{Artefact Research Center}
	\city{Paris}
	\country{France}
}
\email{amaury.de-vitry@artefact.com}

\author{Arthur Garnier}
\affiliation{%
	\institution{Ardian}
	\city{Paris}
	\country{France}
}
\email{arthur.garnier@ardian.com}

\author{Hicham Randrianarivo}
\orcid{0009-0006-1764-5253}
\affiliation{%
	\institution{Artefact Research Center}
	\city{Paris}
	\country{France}
}
\affiliation{%
	\institution{MICS, CentraleSupélec, Université Paris-Saclay}
	\city{Gif-sur-Yvette}
	\country{France}
}
\email{hicham.randrianarivo@artefact.com}

\author{Emmanuel Malherbe}
\orcid{0009-0006-0898-6873}
\affiliation{%
	\institution{Artefact Research Center}
	\city{Paris}
	\country{France}
}
\email{emmanuel.malherbe@artefact.com}

\begin{document}

	\begin{abstract}
		Finance reporting is a natural proving ground for large language models, and the very-long-context capabilities of recent models across all sizes make rigorous evaluation in this domain an increasingly pressing need. Yet most public financial resources reduce the task to plain-text SEC 10-K filings paired with a handful of question--answer items. We release \textsc{Ledger} (Long-context Evaluation of Documents for Grounded Extraction and Retrieval), a corpus of 4{,}999 digitized corporate \emph{annual reports} --- full documents with figures, tables, and narrative, not just regulatory filings.  
        Each report is labeled with 31 consolidated financial KPIs to be extracted and linked to the market's reaction at the earnings date. From this data we derive three evaluation benchmarks spanning the difficulty spectrum: a pure page-level KPI retrieval task with TREC-style relevance judgments over 118{,}048 questions in natural language, a conversational ``needle-in-a-haystack'' single-value lookup, and a full KPI extraction task, both from long, numerically dense reports. We additionally provide human OCR-quality annotations with inter-annotator agreement and the complete extraction, validation, and scoring toolchain. We further demonstrate the dataset's research utility with a case study linking CEO-letter rhetoric to post-publication market impact.
	\end{abstract}


	\maketitle

\section{Introduction}
\label{sec:intro}

Corporate financial reports provide critical data on the health and results of a company, but their significant length and complexity pose a heavy cognitive burden on analysts. Despite these barriers, their release triggers swift, high-volatility market reactions, and the critical window to assess these reports before market opens and reacts is very short, generally of one hour. Consequently, automated tools capable of extracting actionable intelligence from dense financial text in near-real-time would benefit any financial investor.
Large language models (LLMs) promise to read, understand, and extract such information, even more with the rapid growth of context windows -- now reaching hundreds of thousands of tokens even for small, locally hostable models.
Whether models can actually \emph{ground} their answers in such documents is, however, poorly measured. Most public financial resources reduce the problem to short, plain-text excerpts of U.S.\ SEC 10-K filings
paired with a few hundred question--answer pairs~\cite{chen2021finqa,zhu2021tatqa,islam2023financebench}.
Real financial analysis instead operates over the full \emph{annual report}: a long,
visually dense ``glossy'' document interleaving narrative, statutory statements,
hundreds of tables, and figures, where the same quantity (``revenue'') is reported
under company-specific labels, scales, and scopes.

We release \textsc{Ledger}, a resource built to evaluate retrieval and extraction on exactly these documents. We start from 4{,}999 OCR'd annual reports and label them with a
ground-truth KPIs, natural language questions and corresponding page-level relevance.
The resource is organized as a difficulty spectrum---from
locating a page, to extracting one value, to extracting an entire financial statement---and ships with the complete toolchain used to build and score it. Our
contributions are:
\begin{itemize}
  \item \textbf{A corpus of labeled reports}: reports of 738 companies over fiscal years 2009--2024, with 31 consolidated KPIs for each and market data aligned with their earning date. Each of the 118{,}048 KPI labels comes with a natural language question, and page-level relevancy labels (Section~\ref{sec:resource}).
  \item \textbf{Three benchmarks with baselines}: page-level KPI retrieval (TREC qrels), single-value conversational ``needle-in-a-haystack'' lookup, and full multi-KPI extraction, with sparse/learned-sparse IR and four open-weight LLM baselines (Section~\ref{sec:benchmarks}).
  \item \textbf{A research-utility case study} linking CEO-letter rhetoric to
    earnings per share surprise and post-publication market impact (Section~\ref{sec:casestudy}).
  \item \textbf{Open release} under the MIT license for the code and Creative Commons Attribution 4.0 for the data, with a permanent DOI and the
    full extraction/validation toolchain.
\end{itemize}

\section{Related resources}
\label{sec:related}

Finance has been an early and active target for LLMs, both as a training domain
(e.g.\ BloombergGPT~\cite{wu2023bloomberggpt}, trained on proprietary data) and as an
evaluation domain. Numerical reasoning benchmarks such as
FinQA~\cite{chen2021finqa}, ConvFinQA~\cite{chen2022convfinqa} and
TAT-QA~\cite{zhu2021tatqa} pair short table-plus-text snippets with arithmetic
questions; FinanceBench~\cite{islam2023financebench} broadens this to open-book QA
over 10-K filings, and DocFinQA~\cite{reddy2024docfinqa} extends the context to
document scale. Retrieval-oriented work has shown that document structure matters for
financial RAG~\cite{jimenoyepes2024chunking}, and visual document retrieval
benchmarks such as ViDoRe~\cite{loison2026vidore} evaluate page retrieval over
rendered PDFs. \textsc{Ledger} differs along four axes simultaneously:
(i)~it covers \emph{full annual reports}---not just 10-K text---including the glossy
front matter, figures, and CEO letters that 10-Ks omit; (ii)~documents are genuinely
long-context ($\approx$126k tokens, median 124 pages); (iii)~it provides
\emph{authoritative numeric labeling} at scale, with $\approx$30 KPIs per
company-year reconciled against SEC XBRL filings rather than crowd answers; and
(iv)~it couples the documents to a market-reaction signal, enabling downstream
financial-research tasks. The whole resource is openly licensed and accompanied by a
datasheet~\cite{gebru2021datasheets} and FAIR-compliant~\cite{wilkinson2016fair}
metadata.

\section{The \textsc{Ledger} corpus}
\label{sec:resource}

\begin{figure}
    \centering
    \includegraphics[width=0.95\linewidth]{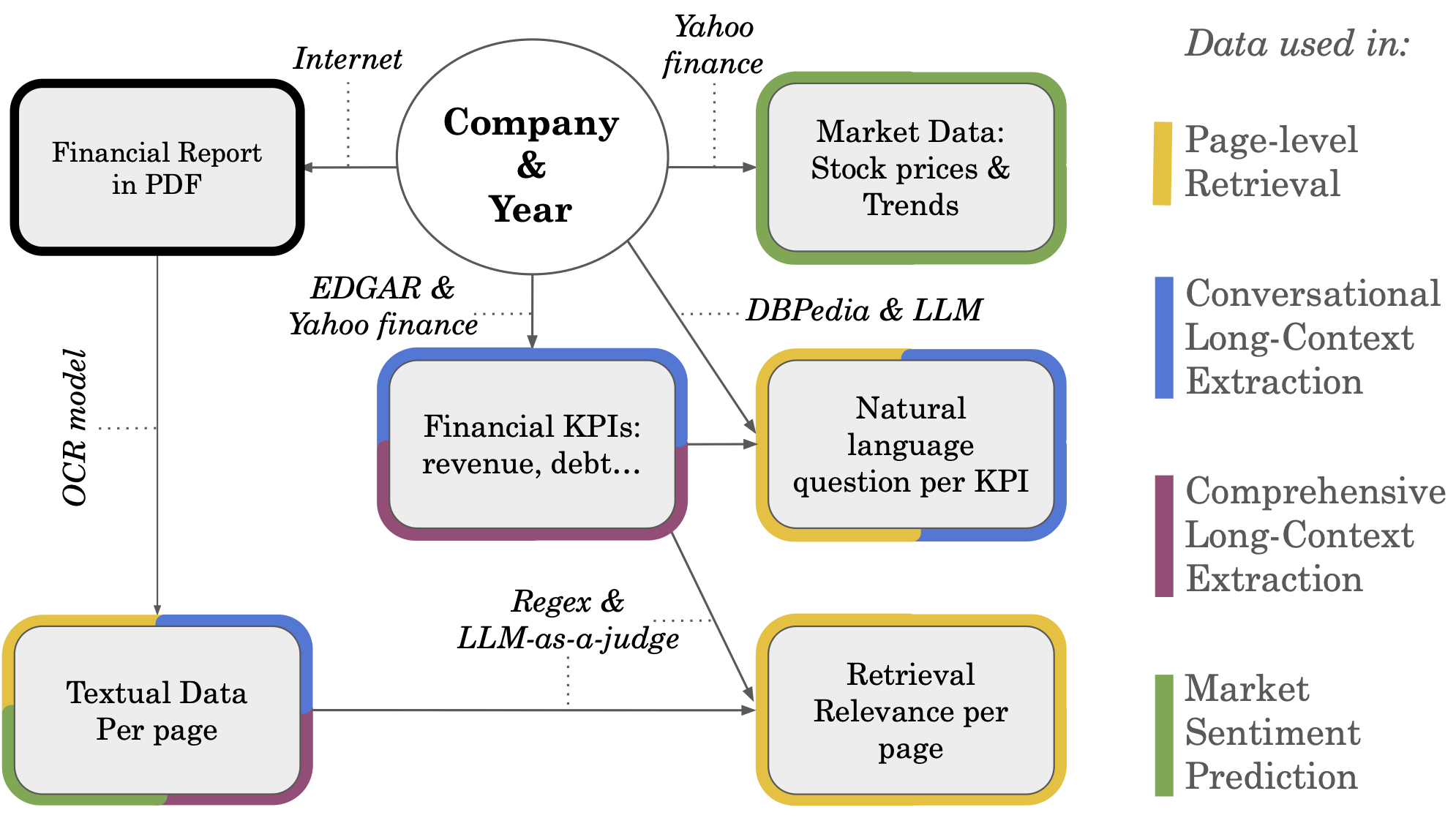}
    \caption{Overview of \textsc{Ledger} corpus.}
    \label{fig:overview}
\end{figure}

\subsection{Financial Reports and OCR}
\textsc{Ledger} corpus is built from a set of 4{,}999 publicly available corporate annual reports in PDF format. Using DeepSeek-OCR-2~\cite{wei2025deepseek, wei2026deepseek}, we converted each \textsc{LEDGER} report into a single page-aligned Markdown file in which pages are delimited by an explicit \texttt{<--- Page Split --->} marker, tables are rendered as HTML/\LaTeX, and per-page raster images are retained for downstream visual tasks. 

Because every downstream task depends on OCR fidelity, we provide a human assessment
of table-extraction quality. Fifteen annotators graded the table alignment and contents $\approx$1{,}150 extracted
financial tables as \texttt{ok}\,/\,\texttt{not\_ok}\,/\,\texttt{uncertain} through a purpose-built interface that renders each detected table next to its source page image. 
A 273-table subset was triple-coded to measure agreement: we observe 87.2\% full agreement, 91.3\% pairwise agreement, and a Fleiss' $\kappa$ of \textbf{0.81} (substantial agreement). At the table level,
81.5\% of tables were graded as correctly aligned, giving a concrete, auditable estimate of OCR table fidelity that downstream users can condition on. Overall, number extraction is nearly perfect despite rare occasional misrepresentations; nevertheless, layout alignments can be suboptimal for complex tables.

\begin{table}[t]
  \caption{The \textsc{Ledger} dataset statistics. The Eval companies are a subset of the Total corpus. Tokens were counted using the \texttt{cl100k\_base} tokenizer.}
  \label{tab:stats}
  \small
  \begin{tabular}{@{}lrrr@{}}
    \toprule
    \textbf{} & \textbf{Eval} & \textbf{Total} & \textbf{Avg./report} \\
    \midrule
    Companies          & 111    & 738     & - \\
    Reports            & 494    & 4,999   & - \\
    Pages              & 47,820 & 691,858 & 138 \\
    Tokens             & 41M    & 631M    & 126,274 \\
    KPIs \& Questions  & 13,519 & 118,048 & 23.6 \\
    Relevant pages (qrels)  & 392,484      & 2,054,279       & - \\
    Daily stock prices     & -      & 800k    & 160 \\
    \bottomrule
  \end{tabular}
\end{table}

\subsection{KPI labeling}
\label{sec:kpis}
We attach consolidated KPIs to every company-year report, organized across the
three financial statements: \emph{(i) income} -- revenue, cost of revenue, gross profit, R\&D, SG\&A, operating income, interest expense, income-tax expense, net income, basic and diluted EPS; \emph{(ii) balance sheet} -- total assets, total liabilities, stockholders' equity and its incl.-NCI variant, cash \& equivalents and its incl.-restricted variant, long-term debt at three scopes, short-term borrowings, inventory, receivables, payables, shares outstanding;
and \emph{(iii) cash flow}
-- operating, investing, financing flows, capex, depreciation \& amortization, dividends paid.
EBITDA is deliberately excluded: it is reporting-standard dependent and derivable from the released components.
Values are routed through a three-tier source waterfall: SEC EDGAR
\texttt{companyfacts} XBRL for U.S.\ listings, Yahoo!Finance API (through the \texttt{yfinance} python package) as a fallback for non-U.S.\
listings, and Alpha Vantage as an opt-in gap-fill that never overwrites the first two.
The result is 118{,}048 audited and consolidated facts with per-KPI yearly coverage above 85\% for the core line items, with a fiscal year convention that handles 52 and 53-week year reportings.

\subsection{Natural language questions and relevant pages for KPI}
For every (company, year, KPI) triplet, we generated a natural-language question to simulate a usage in a chat user interface. To do so, we first collected the company alternative names using DBpedia \cite{auer2007dbpedia}, in order to automatically reproduce the variability of semantics of human users as done in \cite{malherbe2016bridge}. Second, we generated different variants of the question for the given KPI using Gemini 3.1 Pro~\cite{gemini31pro2026}, in the form of templates where \textit{company name} and \textit{year} are to be replaced. In practice, since the KPI and the company occur several times in our set of triplets, we sampled uniformly the KPI question template and company name once for every triplet.

In a context of RAG on the pages of the corresponding report, we annotated the relevant page for each query. To do so, we took the corresponding ground-truth KPI value, and searched for pages of the report containing this value, producing TREC-style relevance judgments~\cite{voorhees2005trec}. 
From 118{,}048 questions we mine over a million candidate (query, page) pairs via unit-normalized value matching and grade each on a 0/1/2 scale (not relevant / contextual mention / primary source) with an LLM judge (in our case Qwen 3.6~\cite{qwen3.6-27b}), yielding a graded \texttt{qrels} file. A manual spot-check of 60 graded pairs by a domain expert yielded 91.6\% agreement with the judge; a full multi-annotator study is planned for a future version of this work.

\subsection{Market-reaction linkage}
For each U.S.\ company-year we link the report to the market's reaction at publication. Using EDGAR's \texttt{submissions} feed we select the original 10-K (excluding amendments) whose filer-labelled FY matches, take the earliest \texttt{acceptanceDateTime} (UTC), and classify it as pre-market / intraday / after-hours against NYSE hours in Eastern time.
We consider raw returns, defined as:
\begin{equation}
\label{eq:raw_return}
r_{h} = \frac{P_t - P_{t+h}}{P_t},
\end{equation}
where $P_t$ is the closing price on the earnings date (or the prior close if published pre-market). 
Note that since we will compare with a past date (i.e. $h<0$), our choice of sign means that $r_h>0$ expresses a rise in the stock price. 
We also compute the Earnings Per Share (EPS) Surprise as
\begin{equation}
\label{eq:eps_surprise}
\text{EPS Surprise} = \frac{\text{Reported EPS} - \text{Consensus EPS}}{|\text{Consensus EPS}|}
\end{equation}
where the denominator ensures the sign of the surprise is preserved, and the EPS is for the fourth quarter of the financial year.

\section{Benchmarks and baselines}
\label{sec:benchmarks}

The resource induces three tasks of increasing difficulty over the same documents and ground truth label: find the page given the question, extract one KPI given the question, extract every KPI.
For the two extraction tasks (Sections~\ref{sec:needle}--\ref{sec:multikpi}) we report the  \textbf{recall} $=$ correct values\,/\,gold values (a non-answer counts as a miss) and \textbf{precision} $=$ correct\,/\,attempted (non-answers and unverifiable extras excluded). 
For the single-KPI extraction, recall coincides with the exact-match \emph{accuracy} reported in the long-context literature~\cite{hsieh2024ruler,liu2024lost}. 
To fit into our 4-GPU setup in a reasonable time, we run our experiments over 494 labeled reports of our corpus, with 13,519 KPI ground-truth labels in total (27.2 KPI/report). For the two first tasks, we kept 10{,}000 questions for which there is at least one relevant page.

\subsection{Page-level retrieval}
\label{sec:retrieval}

With the scenario of a RAG in mind, we first evaluated how a retrieval using the question as a query would retrieve the relevant page from the relevant report: $(\textit{question, report})\to \textit{relevant page}$.
Table~\ref{tab:retrieval} reports this page-level retrieval protocol, comparing lexical BM25~\cite{robertsonProbabilisticRelevanceFramework2009} and the learned-sparse SPLADE~\cite{formal2021splade} against the multi-representation dense retriever ColBERT~\cite{khattab2020colbert}.
ColBERT consistently outperforms both BM25 and SPLADE across all metrics. This superior performance is expected because ColBERT uses token-level late interaction to compute relevance scores from fine-grained token-to-token alignments, rather than compressing the entire document into a single vector.
Locating the right page remains exceptionally difficult: ColBERT achieves an MRR of only 0.475 (Recall@5: 0.370), while SPLADE and BM25 drop to 0.386 (Recall@5: 0.272) and 0.324 (Recall@5: 0.265), respectively, confirming that dense numerical pages heavily defeat off-the-shelf retrievers. Note that to run SPLADE and ColBERT, we use 4 GPUs to parallelize report-level evaluation (one model instance per GPU, with dynamic work-stealing dispatch) and an encoding batch size of 32 for both document pages and queries within each report.
\begin{table}[t]
\centering
\caption{Per-document page retrieval, averaged over 10{,}000 questions (494 reports, Test corpus). Recall@$k$/MRR use binary relevance ($\mathrm{rel}\!\ge\!1$); nDCG uses graded gains (0/1/2). Best per row in \textbf{bold}.}
\label{tab:retrieval}
\begin{tabular}{@{}lccc@{}}
\toprule
\textbf{Metric} & \textbf{BM25} & \textbf{SPLADE} & \textbf{ColBERT} \\
\midrule
Recall@1  & 0.069        & 0.099          & \textbf{0.129} \\
Recall@3  & 0.185        & 0.208          & \textbf{0.279} \\
Recall@5  & 0.265        & 0.272          & \textbf{0.370} \\
MRR       & 0.324        & 0.386          & \textbf{0.475} \\
nDCG@5    & 0.221        & 0.245          & \textbf{0.333} \\
nDCG@10   & 0.280        & 0.283          & \textbf{0.383} \\
\midrule
Runtime (in seconds)   & \textbf{66} & 160          & 869 \\
\bottomrule
\end{tabular}
\end{table}

\subsection{Conversational long-context extraction}
\label{sec:needle}
With the scenario of a conversational chatbot in mind, we evaluated the long-context information extraction given a question and its report: $(\textit{question, report})\to \textit{KPI}$.
To do so, the model receives an entire OCR'd report ($\approx$100k tokens) and extract a single specified KPI as a structured JSON object
(\texttt{found}, \texttt{value}, \texttt{unit\_scale}, \texttt{page}). A response is \emph{matched} within $\pm0.05\%$ of ground truth label.
Because the long document prefix is constant per report and
only the question varies, prefix caching cuts prefill by $\approx$21$\times$, making
full-corpus evaluation tractable on a single GPU server. The left block of
Table~\ref{tab:llm} reports four open-weight models: the strongest
(Qwen3.6-27B) reaches 91.4\% recall at 93.5\% precision, while a model with a systematic unit-scaling error (Nemotron) collapses to 15.0\%.

\subsection{Comprehensive long-context extraction}
\label{sec:multikpi}
The hardest task asks a model to extract all 31 KPIs from a report in a single pass: $\textit{report}\to \textit{all KPIs}$. 
We evaluated this task on ground-truth labels by recall (coverage of true facts) and precision (correctness of
emitted facts). The right block of Table~\ref{tab:llm} shows that single-value extraction capability does \emph{not} transfer: Ministral, second-best at the needle task ($87.9\%$), collapses
to $41.4\%$ recall under structured extraction, emitting many hallucinated cells;
conversely Nemotron, near-useless at the needle task, recovers to $65.5\%$ recall once
schema constraints suppress its scaling error. Qwen3.6-27B is the only model strong on
both, and no model exceeds $80\%$ recall---establishing the task as an open challenge.

\begin{table}[t]
\centering
\caption{LLM baselines for long-context extraction on Test corpus, all as recall (R) and precision (P) with a match tolerance of $\pm0.05\%$.
\emph{Conversational}: 10{,}000 questions.
\emph{Comprehensive}: 494 reports, 13{,}519 cells. Rows ordered by recall; best per column in bold.}
\label{tab:llm}
\small
\setlength{\tabcolsep}{4.5pt}
\begin{tabular}{@{}lccc cc@{}}
\toprule
 & \multicolumn{2}{c}{\textbf{Conversational}} & \multicolumn{2}{c}{\textbf{Comprehensive}} \\
  & \multicolumn{2}{c}{\textbf{(single-KPI)}} & \multicolumn{2}{c}{\textbf{(Multi-KPI)}} \\
\cmidrule(lr){2-3}\cmidrule(lr){4-5}
\textbf{Model} & \textbf{R} & \textbf{P} & \textbf{R} & \textbf{P} \\
\midrule
Qwen3.6-27B~\cite{qwen3.6-27b}          & \textbf{91.4} & \textbf{93.5} & \textbf{77.2} & \textbf{85.7} \\
Ministral-3-14B~\cite{liu2026ministral}      & 87.9 & 88.6 & 41.4 & 43.6 \\
gpt-oss-20b~\cite{openai2025gptoss120bgptoss20bmodel}          & 85.3 & 86.8 & 65.7 & 73.2 \\
Nemotron-3-Nano-30B~\cite{nvidia_nemotron_nano_v3_2025}  & 15.0 & 15.3 & 65.5 & 78.0 \\
\bottomrule
\end{tabular}
\end{table}

\section{Case study: Market Sentiment Prediction}
\label{sec:casestudy}

We investigated the value of our corpus for a more challenging research question: whether the \emph{rhetoric} in the reports carries signal about future fundamentals and market reaction. To do so, we considered six highly liquid industries: specialty chemicals, auto parts, packaged foods, oil \& gas exploration \& production, oil \& gas equipment \& services, and mortgage REITs, spanning over the fiscal years 2017-2022.
On the non-10K reports of these sectors, we decided to evaluate the signal brought solely from the CEO letter.

As models, we compared different backbone encoders, that we applied on the full tokens sequence, added a linear layer atop the averaged embeddings, with an $L_2$ regularization on the linear weights and frozen weights for the backbone model.
We trained the model to predict two targets: whether the return $r_{-90d}$ (Equation~\ref{eq:raw_return}) falls in the top 10\%, the bottom 10\%, and similarly for the EPS Surprise (Equation~\ref{eq:eps_surprise}), each framed as identifying the top/bottom decile (i.e.\ a 10\% prevalence).
Table~\ref{tab:casestudy} reports PR-AUC (5-fold stratified CV) for ten encoders
against a baseline with Multinomial Naive Bayes on a bag-of-words. Several encoder/target combinations land well above
the $0.10$ random baseline---e.g.\ $0.47$ for positive EPS surprises and $0.44$ for
negative 90-day returns---indicating a genuine, if modest, textual signal and
illustrating the kind of cross-modal study (narrative text $\rightarrow$ realized
financials) our corpus enables.
\begin{table}[t]
\centering
\caption{PR-AUC when predicting EPS surprise and $r_{-90d}$ from the CEO-letter (top and low deciles); best per column in \textbf{bold}.}
\label{tab:casestudy}
\small
\setlength{\tabcolsep}{6pt}
\begin{tabular}{lccccc}
\toprule
 & \multicolumn{2}{c}{\textbf{Surprise EPS}} & \multicolumn{2}{c}{\textbf{$r_{-90d}$}} \\
\cmidrule(lr){2-3} \cmidrule(lr){4-5}
\textbf{Encoder} & \textbf{Top 10\%} & \textbf{Low 10\%} & \textbf{Top 10\%} & \textbf{Low 10\%} \\
\midrule
BGE-large-en~\cite{bge_embedding}       & \textbf{0.469} & 0.212          & 0.440          & 0.335          \\
BGE-M3~\cite{bge-m3}      & 0.429          & \textbf{0.389} & \textbf{0.442} & 0.254          \\
ModernBERT \cite{modernbert} & 0.421          & 0.353          & 0.384          & 0.285          \\
Gemma~\cite{vera2025embeddinggemma}      & 0.415          & 0.350          & 0.420          & \textbf{0.403} \\
MiniLM-l6~\cite{reimers-2019-sentence-bert}     & 0.403          & 0.286          & 0.425          & 0.224          \\
MPNet~\cite{song2020mpnet}      & 0.378          & 0.320          & 0.440          & 0.331          \\
MPNet (NLI) & 0.292          & 0.247          & 0.374          & 0.167          \\
MiniLM-l12~\cite{reimers-2019-sentence-bert} & 0.366          & 0.246          & 0.366          & 0.218          \\
RoBERTa \cite{liu2019roberta}    & 0.347          & 0.202          & 0.384          & 0.300          \\
EuroBERT\cite{boizard2025eurobert}   & 0.293          & 0.248          & 0.310          & 0.248          \\
\hline
BoW+Multi NB       & 0.277          & 0.218          & 0.286          & 0.244          \\
\bottomrule
\end{tabular}
\end{table}

\section{Availability, license, and ethics}
\label{sec:availability}

\textbf{Availability.} The corpus, KPI tables, qrels, benchmark splits, model
predictions, and the complete fetch/extract/validate/score toolchain are released under the MIT (for all code productions) license and CC-BY-4.0 (for all data). All datasets are publicly available on HuggingFace (\url{https://huggingface.co/collections/artefactory/ledger}) and all code production is available on our public GitHub repository
(\url{https://github.com/artefactory/LEDGER}). The release includes
datasheets~\cite{gebru2021datasheets} and FAIR-compliant~\cite{wilkinson2016fair} metadata, and
reuses the standard TREC \texttt{qrels} format so existing IR tooling
(e.g.\ \texttt{trec\_eval}) applies directly.

\textbf{Ethics and limitations.} All documents are public regulatory filings and
investor reports; no personal data is involved. KPI ground truth derives from official
XBRL filings and third-party financial APIs, and may contain restatements or vendor
errors; we therefore publish provenance (source and exact tag) for every value and a
human OCR-quality audit (Section~\ref{sec:resource}), but the data is provided for research and is \emph{not} investment advice. OCR introduces errors in dense tables, which the annotation layer quantifies. The benchmark is U.S.-centric and bounded to 2017--2022 for market-reaction data (no EDGAR equivalent for non-U.S.\ listings); we intend to extend both the window and the company universe.

\textbf{Conclusion.} \textsc{Ledger} turns 4,999 real annual reports into a corpus of labeled
documents, evaluation suite spanning retrieval and conversational long-context
extraction. Strong off-the-shelf systems leave clear headroom on every task, and the
released toolchain lets the community extend the resource along the time, industry, country and
modality axes.

\end{document}